\documentclass{article}

\usepackage{amsfonts}
\usepackage{graphicx}
\usepackage{algorithm}
\usepackage{algorithmic}
\usepackage{bbm}
\usepackage{bm}
\usepackage{amsmath}
\usepackage{wrapfig}
\usepackage{subcaption}
\usepackage{caption}
\usepackage[final]{corl_2019} 

\title{Asynchronous Methods for Model-Based Reinforcement Learning}
\author{Yunzhi Zhang$\thanks{Equal contribution}$ \\
UC Berkeley \\
\texttt{yunzhi@berkeley.edu} \\
\And
Ignasi Clavera$^*$ \\
UC Berkeley \\
\texttt{iclavera@berkeley.edu} \\
\And 
Boren Tsai \\
UC Berkeley \\
\And
Pieter Abbeel \\
UC Berkeley \\
}

\begin{document}
\maketitle


\begin{abstract}
Significant progress has been made in the area of model-based reinforcement learning.  State-of-the-art algorithms are now able to match the asymptotic performance of model-free methods while being significantly more data efficient. However, this success has come at a price: state-of-the-art model-based methods require significant computation interleaved with data collection, resulting in run times that take days, even if the amount of agent interaction might be just hours or even minutes.  When considering the goal of learning in real-time on real robots, this means these state-of-the-art model-based algorithms still remain impractical. In this work, we propose an asynchronous framework for model-based reinforcement learning methods that brings down the run time of these algorithms to be just the data collection time.  We evaluate our asynchronous framework on a range of standard MuJoCo benchmarks.  We also evaluate our asynchronous framework on three real-world robotic manipulation tasks. We show how asynchronous learning not only speeds up learning w.r.t wall-clock time through parallelization, but also further reduces the sample complexity of model-based approaches by means of improving the exploration and by means of effectively avoiding the policy overfitting to the deficiencies of learned dynamics models.

\end{abstract}



\vspace{-0.1cm}
\section{Introduction}
\vspace{-0.25cm}
Autonomous skill acquisition has the potential to dramatically expand the tasks robots can perform ranging from manufacturing to household robotics. In real robotic agents, where data gathering is typically expensive, low sample complexity algorithms are required. Model-based reinforcement learning (RL)~\cite{kaelbling1996survey} offers the potential to be data-efficient while achieving the same learning capabilities as model-free RL by first learning a predictive model of the environment and then deriving a controller from it.

In recent years, significant advances have been made in deep model-based reinforcement learning. Model-based algorithms presented in ~\cite{janner2019mbpo, chua2018deep, clavera2018mbmpo, buckman2018steve} achieve the same asymptotic performance as model-free algorithms while requiring an order of magnitude less data. However, these impressive results have been achieved at the cost of increasing the computational burden of model-based algorithms. Tools such as ensembles and probabilistic models have been key ingredients, preventing the policy from overfitting to the deficiencies of the learned model. As a result, while state-of-the-art model-based methods require just a few hours of agent interaction to learn complex tasks, they can nevertheless take days to train. For instance, the algorithm presented in~\cite{clavera2018mbmpo} takes less than three hours of real-world interaction to learn a locomotion behaviour, but the total training time is of 2.2 days~\cite{wang2019mbbench}.
A need for algorithms that are both sample efficient and computationally fast is even more pressing when considering that these algorithms present a large number of sensitive hyperparameters. Altogether, slow experimentation and extensive hyperparameter search constitute a major barrier for the applicability of model-based methods to real-world robotics.

To bring down the wall-clock time of current model-based RL algorithms, we propose an asynchronous strategy where data collection, model learning, and policy improvement take place in parallel and asynchronously. Aside from speed, our method has two further benefits: First, learning the model while training the policy prevents the policy from overfitting to the deficiencies of the model, effectively regularizing the policy learning step~\cite{luo2018algorithmic}. Second, collecting each rollout using the latest policy trained during the policy improvement process diversifies the data collected, which results in better predictive models. 

The main contribution of our work is a general asynchronous framework for model-based reinforcement learning that reduces the run time of current model-based algorithms to be just the sampling time. It achieves better sample complexity than the classical sequential versions, and removes some hard-to-tune hyperparameters in current model-based approaches, such as the number of trajectories to collect or the number of gradients steps to take. Our experimental evaluation illustrates the strengths of our framework on four standard MuJoCo~\cite{2012mujoco} locomotion tasks. For instance, we were able to learn an optimal policy in high dimensional and complex quadrupedal locomotion within 60 minutes, while the classic sequential version takes more than 10 times longer. Finally, we showcase the effectiveness of our approach in real robotic manipulation skills 
that include block stacking and shape matching. In these cases, our asynchronous framework was able to succeed at each of the tasks within 10 minutes of wall-clock time. 
On the real robotic tasks our approach closely matches the performance of prior specialized work for such complex contact manipulation~\cite{levine2015learning}, while being more general.
Code of parallel and sequential implementation of model-based algorithms, as well as videos of our method on thr real robot environment, can be found at our website.\footnote{https://sites.google.com/view/asynch-mb-rl/home}

\vspace{-0.1cm}
\section{Related Work}
\vspace{-0.25cm}
In this section, we discuss related work, including model-based reinforcement learning, asynchronous learning in the context of RL, and finally real robotic learning with RL. 

\textbf{Model-based reinforcement learning.} Model-based reinforcement learning methods are promising candidates for real-world sequential decision-making problems due to their data efficiency~\cite{kaelbling1996survey}. Current model-based RL algorithms generally fall into one of three categories: Dyna-style algorithms, where the model is used to create imaginary experience for a model-free algorithm~\cite{kurutach2018model, clavera2018mbmpo, luo2018algorithmic, janner2019mbpo, sutton1991dyna, sutton1991planning, buckman2018steve}; model predictive control (MPC) algorithms, where the model is used for planning at each time-step~\cite{nagabandi2017neural, chua2018deep}; and policy search with backpropagation-through-time approaches, which exploit the model derivatives~\cite{deisenroth2011pilco, heess2015learning, tassa2012synthesis, levine2013guided}. In this work, we focus on asynchronous versions of Dyna-style approaches~\cite{kurutach2018model, clavera2018mbmpo}. Dyna-style methods learn a parametric policy, which burdens the training process but make the methods able to scale up to high dimensional domains and become suitable for real robotics tasks. In contrast, MPC approaches do not internalize experience into a parametric policy. This results in a faster training process but computationally expensive test time. Moreover, MPC methods tend to scale poorly to high dimensional domains~\cite{wang2019mbbench}, making them often impractical for real-time feedback control.
Nevertheless, our framework can easily be extended to MPC and backpropagation-through-time approaches. ~\citet{wang2019mbbench} give a general overview and comparison of approaches from all three families.

\textbf{Asynchronous Learning.} 
The Hogwild! algorithm~\cite{recht2011hogwild} popularized asynchronous learning by showing that lock-free asynchronous stochastic gradient descent (SGD) is able to out perform its synchronous version.
Later on, \citet{dean2012largescale} demonstrated its benefits when training deep neural networks.
Inspired by this, \citet{nair2015gorila} were the first to apply asynchronous training to deep reinforcement learning. And further work has extended these results to more efficient algorithms~\cite{Babaeizadeh2016ga3c, espeholt2018impala, hess2017dppo, mnih2016asynchronous, stooke2018accelrl}.
%
However, previous work has focused on model-free algorithms~\cite{mnih2015human, ppo, precup2000eltraces, precup2001tdlearning}, where large amounts of data are required, and distributed data collection is crucial for fast learning. In the case of real robotics agents, however, parallel data collection requires having multiple robots, which can easily be prohibitively expensive~\cite{gu2016asynchrobotics}. 
In our asynchronous framework parallelization occurs across the different phases of model-based RL algorithms, rather than across multiple agents collaborating on one single phase such as collecting experience. 




\textbf{Real Robot Learning.} Prior work on model-based reinforcement learning on real robotic agents has explored a diversity of schemes for dynamics learning, including Gaussian Processes~\cite{deisenroth2011pilco}, mixture models~\cite{moldovan2015optimsm}, and local linear models~\cite{lioutikov2014sample}. In this work we focus on learning dynamics model parametrized by deep neural networks, which offer the potential to scale up to higher dimensional domains and more complex tasks. While deep dynamics models has been previously used on real robots~\cite{nagabandi2018l2a}, it has been done using MPC type approaches. Another line of work has applied pure model-free RL~\cite{ haarnoja2018sacapp, haaronja2018l2walk, hafner2007rlrobcup, gullapalli1992learning}. For instance, ~\citet{gu2016asynchrobotics} used an asynchronous data collection method for door opening. However, model-free RL is still significantly more data inefficient than model-based methods, which hinders its applicability to general real robotics learning. The real robotic tasks attempted in this work are similar to the ones proposed by ~\citet{levine2015learning}. In that work, however, they use a specialized method for contact rich manipulation tasks.


\vspace{-0.1cm}
\section{Model-Based Reinforcement Learning}
\vspace{-0.25cm}
A discrete-time finite Markov decision process (MDP) $\mathcal{M}$ is defined by the tuple $(\mathcal{S}, \mathcal{A}, p, r, \gamma, p_0, H)$. 
Here, $\mathcal{S}$ is the set of states, $\mathcal{A}$ the action space, $p(s_{t+1}|s_t, a_t)$ the transition distribution,
$r: \mathcal{S} \times \mathcal{A} \rightarrow \mathbb{R}$ is a reward function, $p_0: \mathcal{S} \to \mathbb{R}_+$ represents the initial state distribution, $\gamma$ the discount factor, and $H$ is the horizon of the process. We define the return as the sum of rewards $r(s_t, a_t)$ along a trajectory $\tau := (s_{0}, a_{0}, ..., s_{H-1}, a_{H-1}, s_{H})$. The goal of reinforcement learning is to find a policy $\pi: \mathcal{S} \times \mathcal{A} \rightarrow \mathbb{R}^+$ that maximizes the expected return, i.e.:
    $\max_{\pi} J(\pi) = \mathbb{E}_{\begin{subarray}{l}{
a_t \sim \pi}\\
s_{t} \sim p
\end{subarray}}
[\sum_{t=1}^{H}\gamma^{t} r(s_t, a_t)] $.

Model-based RL methods learn the transition distribution, also known as dynamics model, from the observed transitions. This can be done with a parametric function approximator $\hat p_{\bm{\phi}}(s'| s,  a)$. In such case, the parameters $\bm{\phi}$ of the dynamics model are optimized to maximize the log-likelihood of the state transition distribution. Current model-based algorithms often learn an ensemble of models $\{ \hat p_{\bm{\phi}_1}( s'| s,  a), ...,  \hat p_{\bm{\phi}_K}(s'| s,  a) \}$, in which case we denote as $\hat p_{\bm{\phi}}( s'| s,  a)$ the model obtained from having a uniform prior on the ensemble, i.e., $s' \sim p_{\bm{\phi}_I}( s'| s, a)$ where $I \sim \mathcal{U}([K])$.

\vspace{-0.1cm}
\section{Asynchronous Methods for Model-Based Reinforcement Learning}
\vspace{-0.25cm}

\begin{figure*}[t!]
    \centering
    \begin{subfigure}[t]{0.475\textwidth}
        \centering
        \includegraphics[trim={0.2cm 0 0.2cm 0},clip, width=\textwidth]{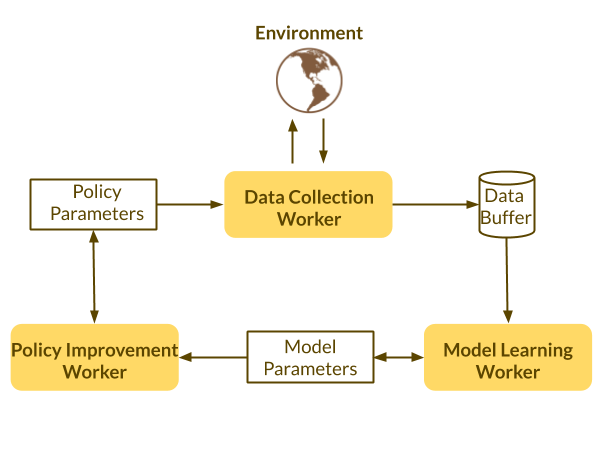}
        \caption{Our proposed asynchronous model-based framework with three workers, communicating exclusively through three servers. The workers do not proceed in a specified order nor wait for the others to complete to execute their own function.}
        \label{fig:asynch_framework}
    \end{subfigure}
    \hspace{0.25cm}
    \begin{subfigure}[t]{0.475\textwidth}
        \centering
        \includegraphics[trim={0.2cm 0 0.2cm 0}, clip, width=\textwidth]{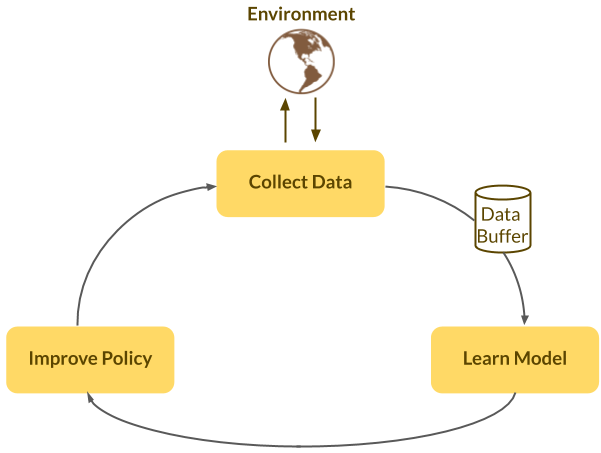}
        \caption{Classic synchronous model-based methods, where three main steps proceed in well-defined order. Each of the steps does not starts running until the preceding one has finalized.}
    \end{subfigure}%
    \vspace{-0.5cm}
\end{figure*}


Typically, model-based algorithms iterate through three phases till convergence: gathering data by interacting with the environment, learning a dynamics model using the gathered data, and improving policy using the learned dynamics model. Previous model-based RL work has made significant strides in decreasing sample complexity. It reduces interaction time with the environment, but shifts more computational load into learning distribution with models to capture uncertainty~\cite{chua2018deep}, and also into learning robust policies~\cite{luo2018algorithmic} or adaptive ones~\cite{clavera2018mbmpo}. As a result, the wall-clock time of running such methods has significantly increased; for instance, training for $200$k timesteps in the Ant environment takes $55$ hours for MB-MPO~\cite{wang2019mbbench}.

Our asynchronous framework, shown in Figure~\ref{fig:asynch_framework},
overcomes this deficiency and further improves sample efficiency of current model-based methods. In the following, we present the general recipe for asynchronous model-based reinforcement learning.

Within the framework, three main tasks of model-based algorithms are assigned to three parallel, independent workers that are respectively dedicated to data collection, model learning and policy improvement. The main task for each worker contains only the minimum amount of work (e.g. collecting one rollout, updating for one epoch or one gradient step). As a result, each worker fetches updates from servers with high frequency and acts in fully asynchronous behavior. Each worker executes three operations: 

\begin{itemize}
\item \textbf{Pull.} Worker gets an update from one of the three servers. For example, for the data collection worker, it pulls the latest policy parameters from the corresponding server.
\item \textbf{Step.} The step operation corresponds to the main function of the worker. For the data collection worker, Step corresponds to collecting one rollout under its local copy of the policy. In the following subsections we explain in detail Step operation for each of the workers.
\item \textbf{Push.} This operation sends the latest parameters or data to one of the three servers. Again, in the case the data collection worker, Push corresponds to pushing the collected rollout to the data buffer.
\end{itemize}

Each worker first checks one specific server either to fetch the latest parameters or to move all data from the remote server to its local buffer. Then it carries out its own step operation, and finally pushes the local change onto another specific server. Each worker loops through this process until a global stopping criterion is met. In the experiments, Section~\ref{sec:exps}, the stopping criterion is set to be a total number of collected trajectories.  

\begin{figure*}
\noindent
\begin{minipage}[t]{0.30\textwidth}
\begin{algorithm}[H]
\begin{algorithmic}[1]
\FOR{$i = 1, ...$}
\STATE Pull policy parameters $\bm{\theta}$
\STATE Collect one trajectory ${(s_t, a_t, s_{t+1})}_t$ with $\pi_\theta$ in the real environment
\STATE Push data $\{ (s_t, a_t, s_{t+1}) \}_t$
\ENDFOR
\end{algorithmic}
\caption{Data Collection}
\label{alg:metatrain}
\end{algorithm}
\end{minipage}
\noindent
\hspace{0.05cm}
\begin{minipage}[t]{0.32\textwidth}
\begin{algorithm}[H]
\begin{algorithmic}[1]
\STATE $\mathcal{D} = \emptyset$
\FOR{$i = 1, ...$}
\STATE Pull samples $\{ s_t, a_t \}_t$
\STATE $\mathcal{D} \leftarrow \mathcal{D} \cup \{(s_t, a_t, s_{t+1}) \}_t$
\STATE Train dynamics model $\hat{p}_{\bm{\phi}}$ for one epoch on $\mathcal{D}$
\STATE Push dynamcis model parameters $\bm{\phi}$
\ENDFOR
\end{algorithmic}
\caption{Model Learning}
\label{alg:metatrain}
\end{algorithm}
\end{minipage}
\noindent
\hspace{0.05cm}
\begin{minipage}[t]{0.35\textwidth}
\begin{algorithm}[H]
\begin{algorithmic}[1]
\STATE Randomly initialize $\bm{\theta}$
\FOR{$i = 1, ...$}
\STATE Pull model parameters $\bm{\phi}$
\STATE Collect imagined samples using $\pi_{\theta}$
\STATE Train policy for one gradient step
\STATE Push policy \\ parameters $\bm{\theta}$
\ENDFOR
\end{algorithmic}
\caption{Policy Improvement}
\label{alg:metatrain}
\end{algorithm}
\end{minipage}
\noindent
\end{figure*}

\textbf{Data collection.} The data collection worker first pulls policy parameters $\theta$ from the server. With the latest policy it proceeds to the step operation, namely collecting one trajectory $\tau = (s_{0}, a_{0}, ..., s_{H-1}, a_{H-1}, s_{H})$. Finally, it pushes the trajectory onto the data buffer and starts over from pulling again. 

\textbf{Model learning.} In each iteration, this worker moves all trajectories from the remote data buffer, if it is not empty, to its local buffer. The local buffer is of fixed size and first-in-first-out. Then, it fits the model for one epoch on the local data buffer. Lastly it pushes model parameter $\bm{\phi}$ to the model parameter server. Since in practice the data collection worker obtains samples at a slower pace than model training, we apply early stopping via computing validation loss on held-out samples. The training of the model stops if the an exponentially moving average of the validation loss increases after an epoch. When new samples are available, the worker resets the rolling average and starts training again. For long-horizon or low-data-frequency tasks where data collection is slow, early stopping is crucial to prevent overfitting.

\textbf{Policy improvement.} In each iteration, the policy improvement worker first pulls from the model parameter server. Then it carries out the specific policy improvement step specified by a model-based algorithm. For instance, in the case of model-ensemble trust-region policy optimization (ME-TRPO)~\cite{kurutach2018model}, this step corresponds to sampling a batch of imaginary trajectories followed by a TRPO update. Finally the worker pushes the improved policy weights $\bm{\theta}$ to the policy parameter server.

Asynchronous learning offers several advantages over sequential learning. First, since the three main processes run in parallel, the running time of the algorithms is reduced to be the total sampling time. Second, since the policy is being learned while collecting data, at the beginning of each rollout a new policy is usually available to the data collection worker, resulting in more diverse data. Third, since the model and policy are learning concurrently, at each policy improvement step a new model is readily available for the policy to fit on. It prevents the policy from overfitting to the model deficiencies, similarly observed in ~\cite{luo2018algorithmic}. Finally, there is no need to specify crucial hyper-parameters for proper learning: number of environment rollouts, number of model epochs, or number of policy gradient steps per iteration.
\vspace{-0.1cm}
\section{Experiments}
\vspace{-0.25cm}
\label{sec:exps}
Here, we will empirically corroborate the claims in the previous sections. Specifically, the experiments are designed to address the following questions: (1) How does the learning speed of our asynchronous framework compare against sequential and model-free baselines? (2) Does asynchronous learning effectively prevent model-bias by regularizing the policy improvement step? (3) Is asynchronous data collection more effective than batch data collection? (4) Is our framework able to rapidly learn complex, real-world manipulation tasks? (5) Is our asynchronous framework brittle to data collection frequency?

To answer the posed questions, we will first evaluate our framework on four continuous control benchmark tasks in the Mujoco simulator~\cite{2012mujoco, gym}. Then, we will analyze its benefits in further depth on a subset of those tasks. And finally, we will showcase its performance on several contact rich object manipulation tasks on the PR2 robot, Figure~\ref{fig:pr2_tasks}. The performance on all the simulated results is averaged over 4 random seeds.

\subsection{Wall-Clock Time Speed-up and Sample Efficiency}\label{sec:exps_time_sample}
We adapt our asynchronous framework to three different model-based algorithms, namely model-ensemble trust-region policy optimization (ME-TRPO)~\cite{kurutach2018model}, a variant of it using proximal policy optimization (PPO)~\cite{ppo} which will refer to as ME-PPO, and model-based meta-policy optimization (MB-MPO)~\cite{clavera2018mbmpo}. We directly compare the performance of the sequential and the asynchronous version of each method, as well as two model-free methods TRPO~\cite{trpo} and PPO~\cite{ppo}. In order to simulate real-world robot experiments, where data-collection is typically the time bottleneck for RL algorithms, we report the wall-clock time that those algorithms would have taken if they were to be run in the real-world. All the experiments have a maximum path length of 200 timesteps. Hence, the time $T$ to collect one trajectory corresponds to 200 times the control frequency, which is an attribute of the environment~\cite{gym}. In the asynchronous case, since data simulation is typically much faster than real-time, the worker responsible for data collection sleeps until the time $T$ elapses, and then starts the next step. 

\begin{figure}[H]
  \centering
  \includegraphics[trim={7.75cm 0.5cm 7.75cm 6.75cm},clip, width=1\textwidth]{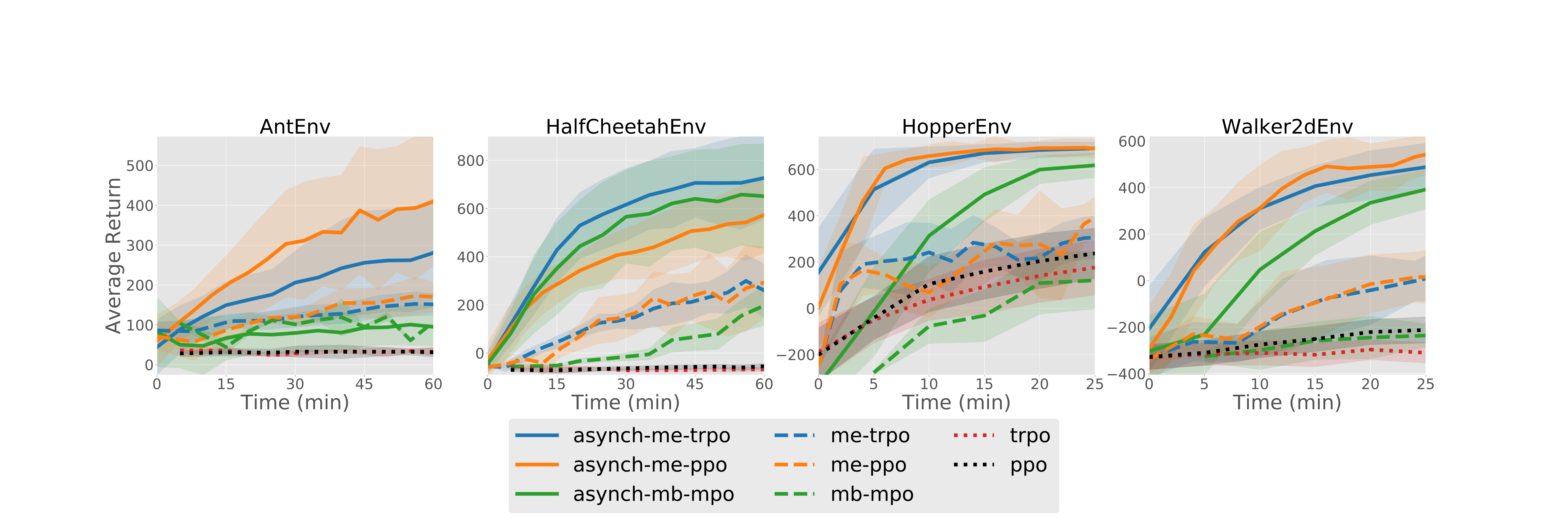}
  \caption{\small{Wall-clock time comparison between asynchronous model-based (solid), synchronous model-based (dashed), and model-free (dotted) methods. Solid lines refer to algorithms within our asynchronous framework, and dashed to its corresponding sequential version. Asynchronous learning significantly speeds up the training time of current model-based algorithms. Best viewed in color.}}
    \label{fig:time}
    \vspace{-0.5cm}
\end{figure}

Figure~\ref{fig:time} shows the performance of the different algorithms in terms of wall-clock time. Here, we see that the asynchronous adaptations significantly speed up the training process. In general, asynch-ME-TRPO and asynch-ME-PPO converge faster than asynch-MB-MPO, and similar relative convergence speed is observed in their synchronous versions.


\begin{figure}[H]
  \centering
  \includegraphics[trim={7.75cm 0.5cm 7.75cm 6.75cm}, clip, width=\textwidth]{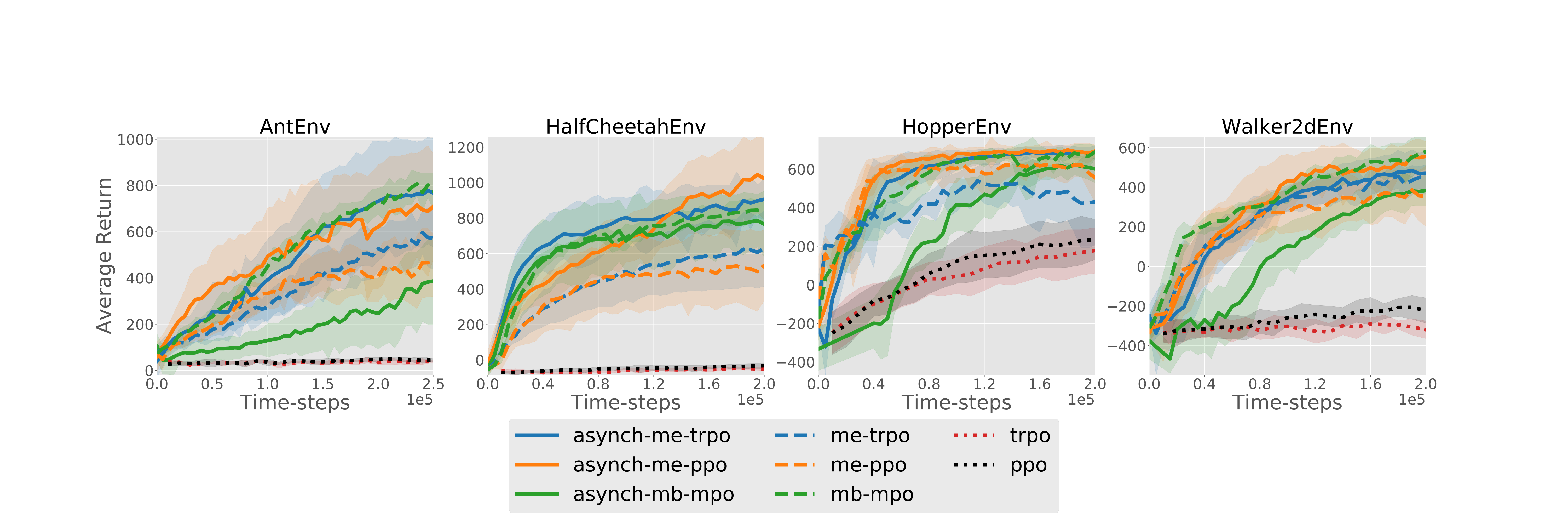}
  \caption{\small{Sample complexity comparison between asynchronous model-based (solid), synchronous model-based (dashed), and model-free (dotted) methods. Solid lines refer to algorithms within our asynchronous framework, and dashed lines refer to corresponding sequential versions. Asynchronous learning in general offers better sample complexity than sequential synchronous learning. Best viewed in color.}}
    \label{fig:samples}
    \vspace{-0.5cm}
\end{figure}

Figure~\ref{fig:samples} shows the performance of the different algorithms in terms of sample complexity. These results show that in general, asynchronous learning converges to its optimal solution faster than their corresponding synchronous methods. They suggest that the asynchronous framework enhances current model-based methods by further reducing the sample complexity of these already data-efficient algorithms.

\subsection{Interleaved Policy Learning and Model Learning}
One aspect that differentiates the asynchronous framework from the synchronous one is that policy updates are interleaved with model updates, whereas in the latter case, the policy does not start taking gradient steps until all the models in the ensemble have either early stopped or reached a pre-determined maximum number of epochs. This section aims to show that such difference benefits the overall learning through policy regularization.

To remove confounding effects, we implemented a partially-asynchronous ME-TRPO
, with each iteration containing two phases: First, collect $N$ rollouts from environment; and second, alternatively fit the model ensemble for $E$ epochs on current dataset and train the policy for $G$ gradient steps with the updated models. The first phase inherits the implementation of synch-ME-TRPO, while the second phase mimics the asynchronous effect by updating the policy with the model parameters before the models are fully trained on the available dataset.

\begin{figure*}[h!]
    \centering
    \begin{subfigure}[t]{0.475\textwidth}
        \centering
        \includegraphics[trim={3.8cm 0.5cm 5cm 2.cm}, clip, width=\textwidth]{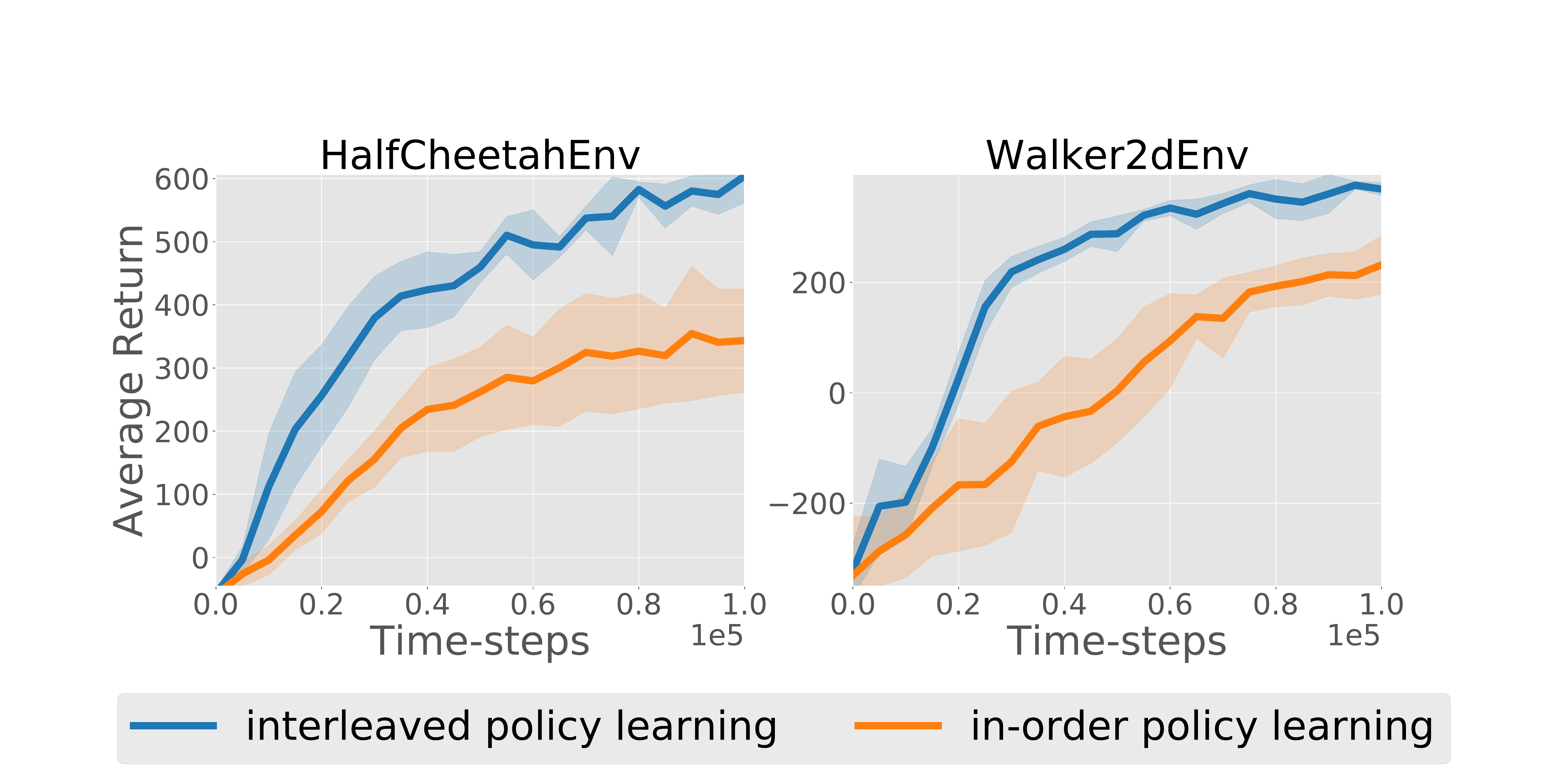}
        \caption{\small{Comparison between interleaved and in-order model and policy updates. Interleaved model and policy updates, which mimics the asynchronous training, leads to better sample complexity by effectively regularizing the policy improvement steps.}}
        \label{fig:regularization}
    \end{subfigure}
    \hspace{0.25cm}
    \begin{subfigure}[t]{0.475\textwidth}
        \centering
        \includegraphics[trim={3.8cm 0.5cm 5cm 2.cm}, clip, width=\textwidth]{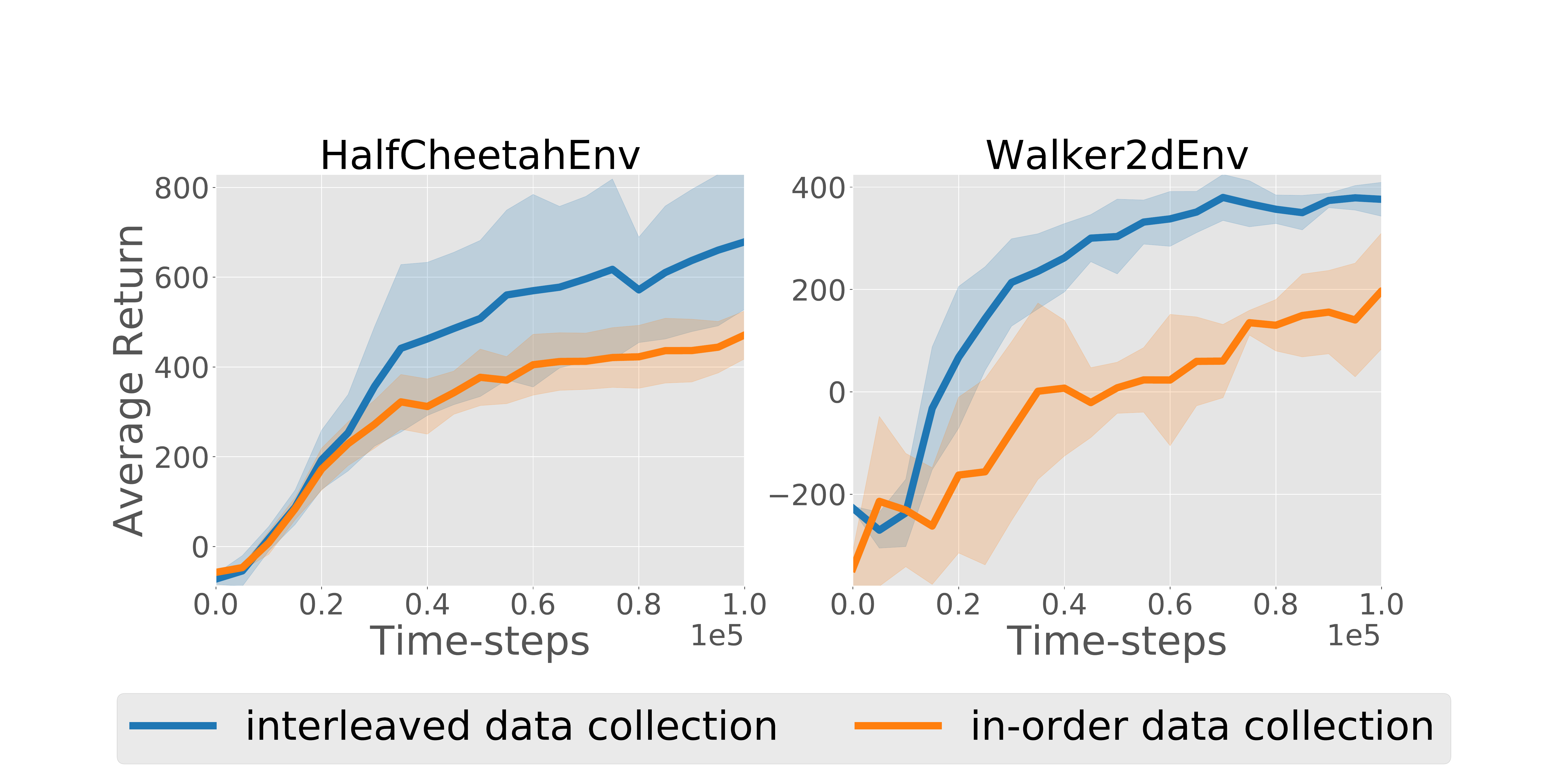}
        \caption{\small{Comparison between interleaved and in-order policy updates and sample collection.  Interleaved policy updates and data collections, which mimics the set up of asynchronous training, reduces sample complexity by collecting more diverse data.}}
         \label{fig:exploration}
    \end{subfigure}%
    \caption{\small{Effects of asynchronous training in learning performance and sample efficiency.}}
\end{figure*}

To test our hypothesis, we compare the aforementioned methods in two Mujoco environments, HalfCheetah and Walker2d. Figure~\ref{fig:regularization} shows that the partially-asynchronous method achieves better sample-efficiency than the synchronous one. It suggests that interleaving model and policy updates, as is the case with the asynchronous framework, helps prevent the policy from overfitting to the model deficiencies.

\subsection{Interleaved Policy Learning and Data Collection}
A second aspect that distinguishes asynchronous methods is that policy learning and data collection are interleaved. That is, environment trajectories are potentially collected under policy even before the policy learner has taken sufficient gradient steps to fit to the current model. 

To investigate whether such a difference improves exploration for data collection, we implemented a second partially-asynchronous ME-TRPO. After acquiring an initial dataset, the trainer loops with two phases: First, fit the model to the obtained dataset; second, alternatively take $G$ policy gradient steps and append a new sampled rollout to the dataset, for a total of $N$ times.

This implementation is compared with synchronous ME-TRPO on HalfCheetah and Walker2d environments, as shown in Figure \ref{fig:exploration}. The result shows the advantage of asynchronous training in terms of sample-efficiency. It suggests that an asynchronous framework effectively encourages data exploration which results in learning benefits.

\subsection{Early Stopping \& Sampling Speed Effect}
In this section, we first investigate the importance of integrating an early-stopping mechanism into our framework. In our framework, we stop training the model if the current validation loss is larger than the exponentially moving average of it. Second, we analyze the effect of the sampling frequency in the data efficiency of asynchronous model-based RL. We analyze these two aspects in asynchronous ME-TRPO in the environments of HalfCheetah and Walker2d.

\begin{figure*}[h!]
    \centering
    \begin{subfigure}[t]{0.475\textwidth}
        \centering
        \includegraphics[trim={3.8cm 0.5cm 5cm 2.5cm}, clip, width=\textwidth]{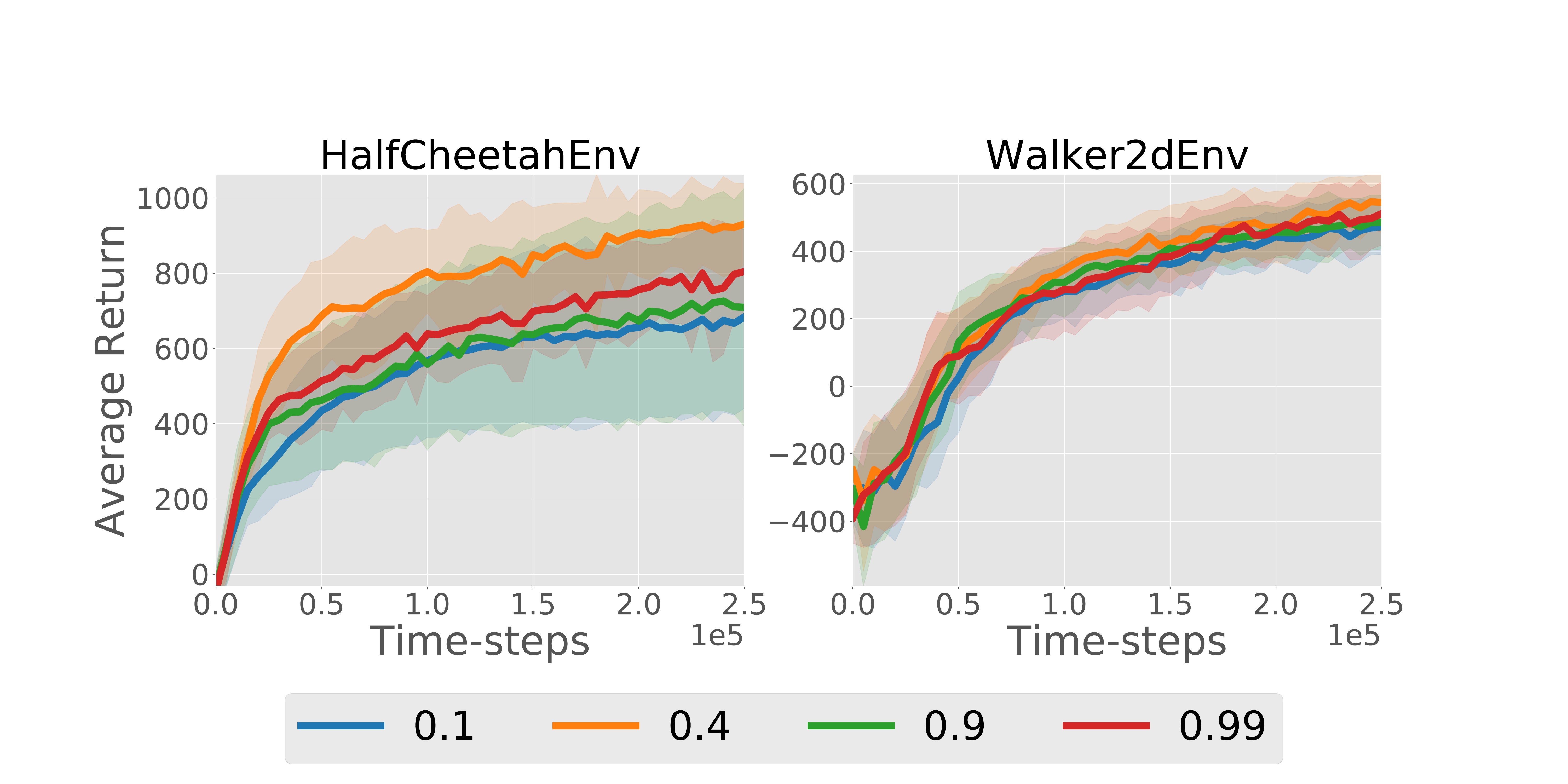}
        \caption{\small{Effect of the early stopping. Lower values weight values in the exponential moving average result in a more aggressive early stopping.}}
        \label{fig:ablations_persistency}
    \end{subfigure}
    \hspace{0.25cm}
    \begin{subfigure}[t]{0.475\textwidth}
        \centering
        \includegraphics[trim={3.8cm 0.5cm 5cm 2.5cm}, clip, width=\textwidth]{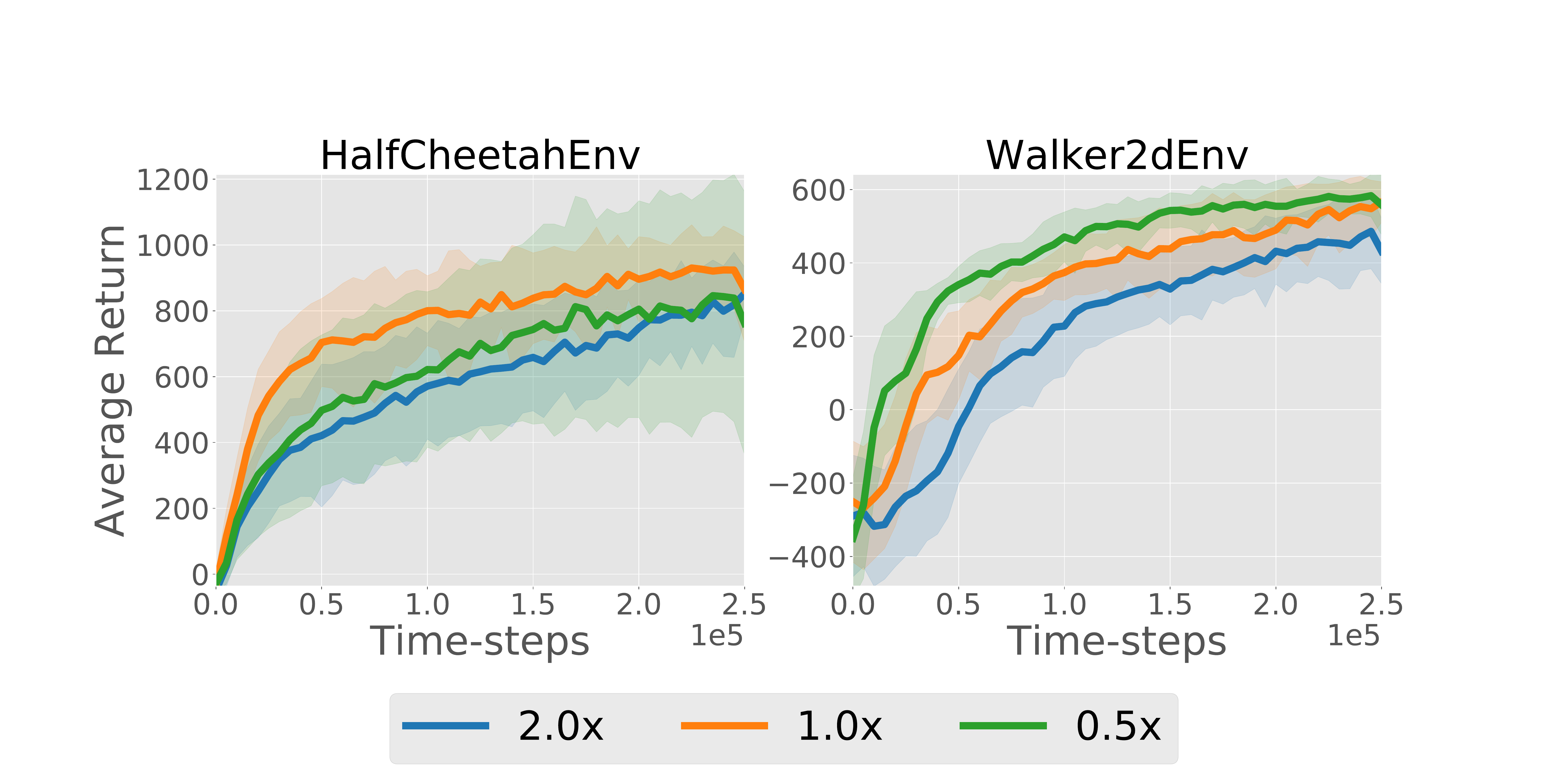}
        \caption{\small{Effect of the sampling speed. Slower data collection benefits the performance in the asynchronous framework by allowing more training on the model and the policy.}}
        \label{fig:ablations_time}
    \end{subfigure}%
    \caption{\small{Early stopping and sampling speed effect on the average returns.}}
\end{figure*}

In Figure~\ref{fig:ablations_persistency} we show how different values of the weight in the exponentially moving average affect performance. In Walker2d, the framework is robust to different degree of early stopping. However, in the HalfCheetah environment, an appropriate value of early stopping leads to faster learning.

The length of a rollout is typically a characteristic of the problem, and thus not tunnable. Here, we compare the performance of our asynchronous training when data collection is carried out at different speeds; specifically, twice the speed and half of it, Figure~\ref{fig:ablations_time}. The result shows that slower data collection, rather than faster, typically leads to better results. We attribute this to the fact that in the asynchronous framework, data collection speed determines the number of gradient steps taken on the model and policy. Slower data collection allows for more model and policy training. Hence in algorithms where model or policy training is particularly slow, the asynchronous framework would benefit from preventing the data collection worker to gather an excessive number of samples. A particular instance of this effect is the increase on sample complexity of asynch-MB-MPO in Section~\ref{sec:exps_time_sample}.

\subsection{Real-World Experiments}
\begin{figure*}[t!]
    \centering
    \begin{subfigure}[t]{0.28\textwidth}
        \includegraphics[trim={3.8cm 0.5cm 3.8cm 2.5cm}, clip, width=\textwidth]{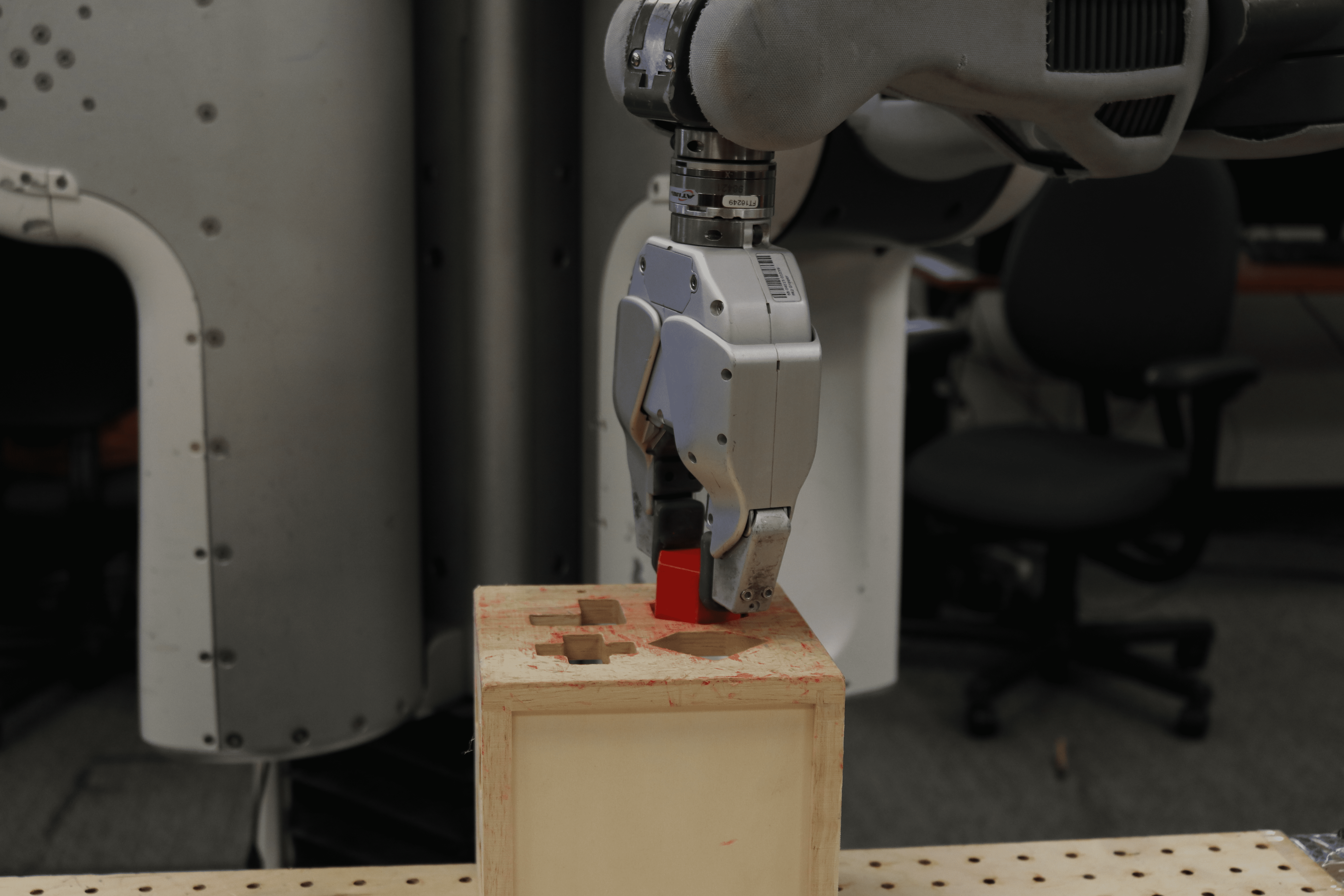}
        \caption{\small{Shape matching, the PR2 robot insert shape into its matching hole.}}
        \label{fig:pr2_shape}
    \end{subfigure}
    \hspace{0.2cm}
    \begin{subfigure}[t]{0.27\textwidth}
        \includegraphics[trim={5cm 0.5cm 3.5cm 3.cm}, clip, width=\textwidth]{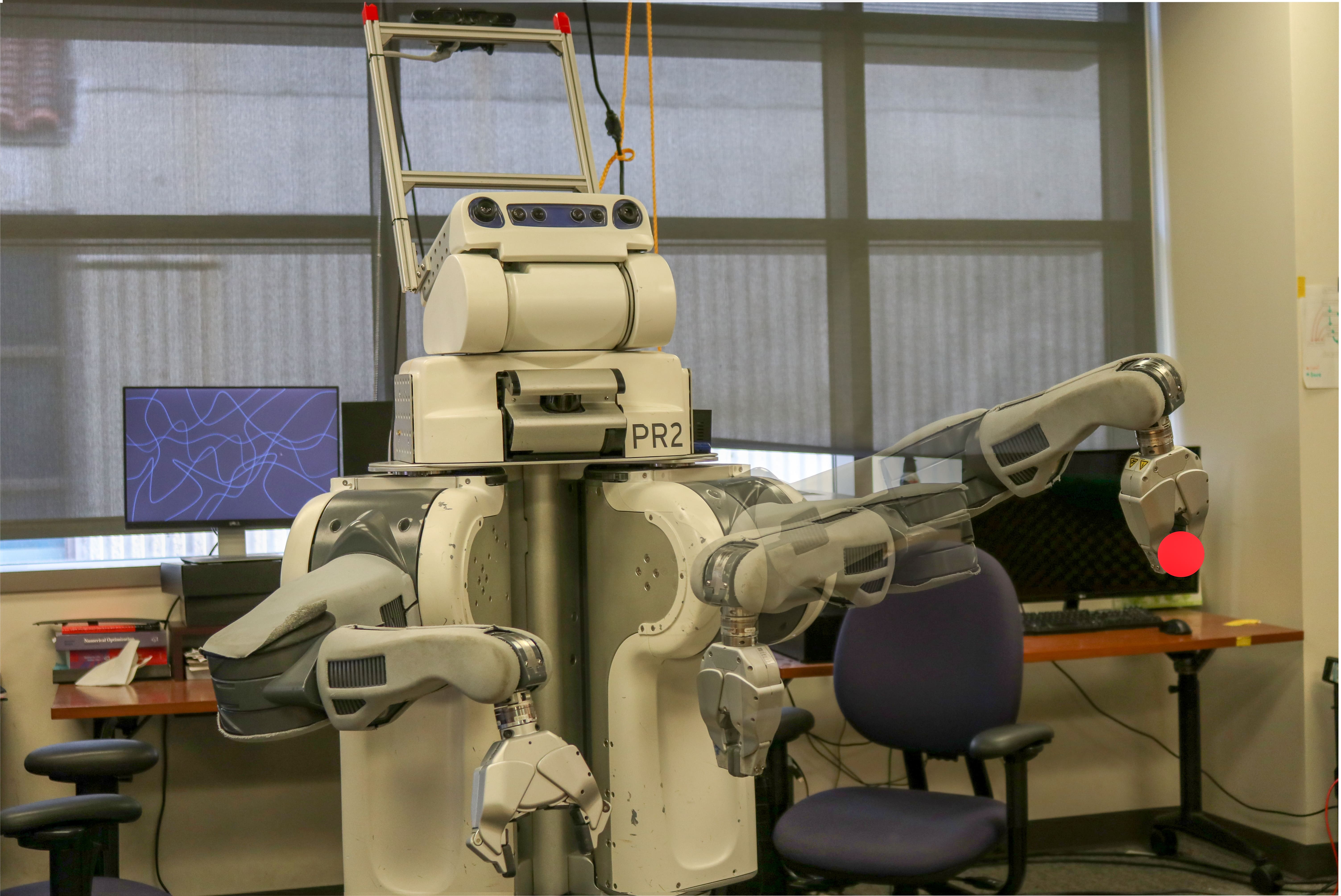}
        \caption{\small{Reaching, the PR2 moves its end-effector to an pre-specified goal.}}
        \label{fig:pr2_reaching}
    \end{subfigure}
    \hspace{0.2cm}
    \begin{subfigure}[t]{0.28\textwidth}
        \includegraphics[trim={3cm 0.5cm 3cm 3cm}, clip, width=\textwidth]{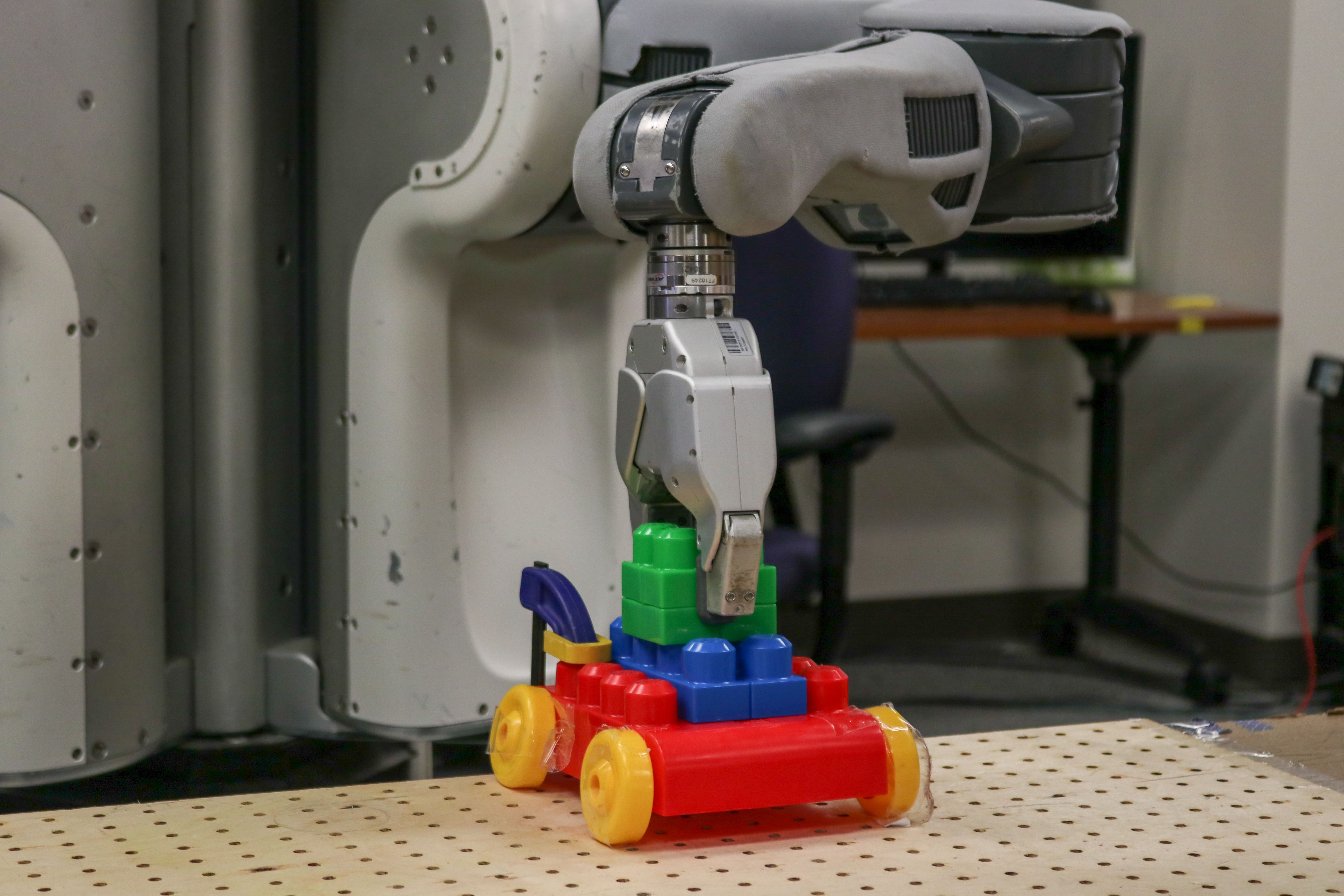}
        \caption{\small{Lego stacking, the PR2 assembles one lego block on top of another.}}
        \label{fig:pr2_lego}
        \end{subfigure}
    \caption{\small{Tasks in our PR2 experiments.}}
    \label{fig:pr2_tasks}
    \vspace{-0.5cm}
\end{figure*}

To best evaluate the real-life efficacy of our proposed asynchronous methods, we expanded our framework's domain of application to include a physical PR2 robot. In particular, we evaluated asynch-MB-MPO in three tasks: reaching a position, inserting a unique shape into its matching hole in a box, and stacking a modular block onto a fixed base.
In the latter two experiments, the manipulated objects were assumed to be fixed extensions of the end-effector, allowing us to use forward kinematics to compute the object's current position. These tasks can be observed in Figures~\ref{fig:pr2_shape}-c. 

The robot itself has a 23-dimensional state space composed of measurements from the active left arm: seven joint angles, seven joint velocities, and nine Cartesian points of the end-effector that determine the pose of the object. The action space was directly the torque commands for the 7-DOF arm. The actions were applied at a frequency of 10 Hz.

The reward function for each task was applied on the distance, $d$, between the current position of the end-effector and a fixed target position. As in~\cite{levine2015learning}, we formulated reward as a mixture of a quadratic penalty and a Lorentzian $\rho$-functions, i.e., $r(d) = -\omega d^{2} - v \log(d^{2} + \alpha)$, where we set $\omega = 1.0$, $v = 1.0$, and $\alpha = 10^{-5}$. The shape of this cost function ensures quick and precise execution of tasks. Based on this reward, we also introduced two scaled quadratic penalties on the magnitude of the joint velocities and applied torques controls to best secure a smooth performance in task completion.

\begin{figure*}[h!]
    \centering
        \includegraphics[trim={5cm 2.5cm 5cm 2.5cm}, clip, width=0.3\textwidth]{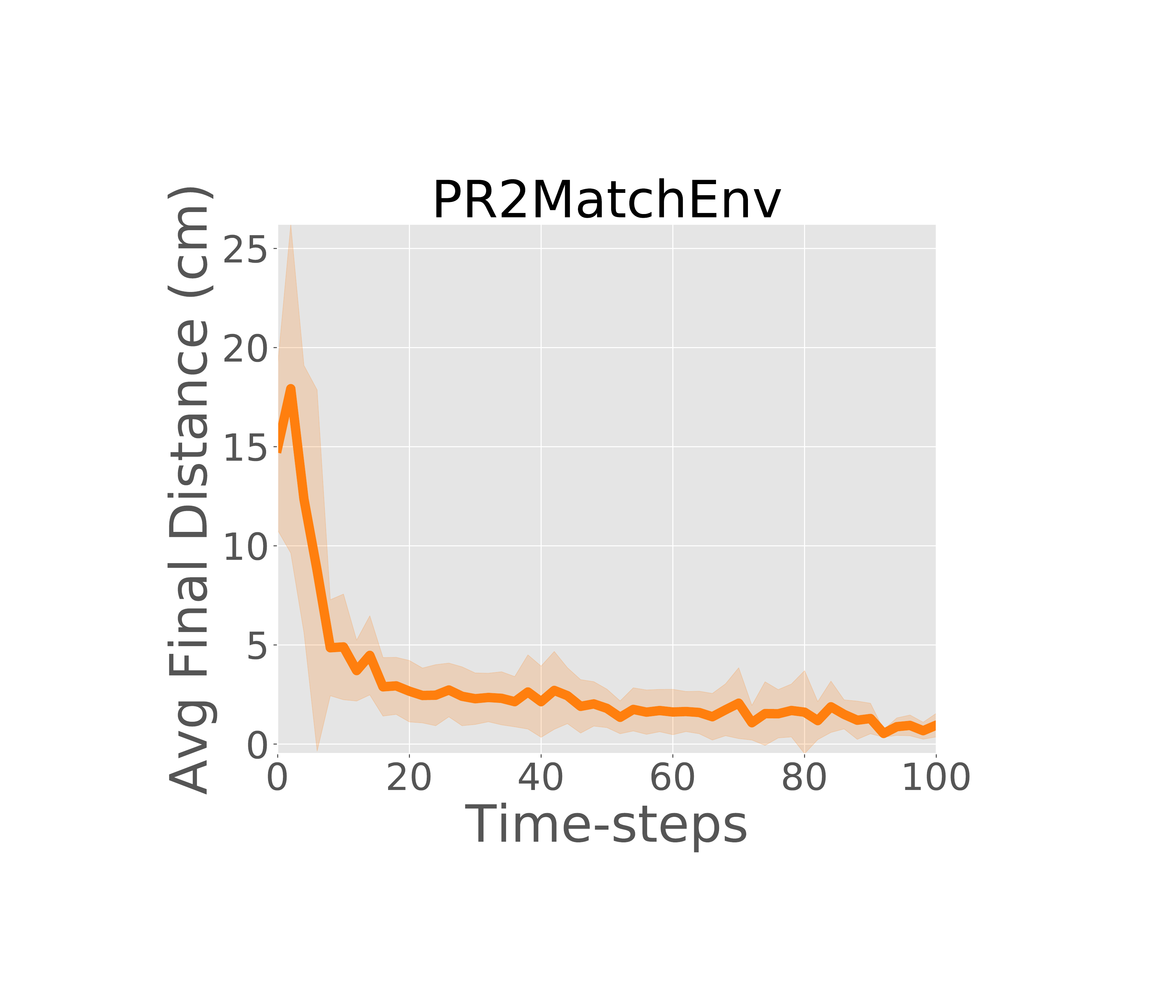}
        \centering
        \includegraphics[trim={5cm 2.5cm 5cm 2.5cm}, clip, width=0.3\textwidth]{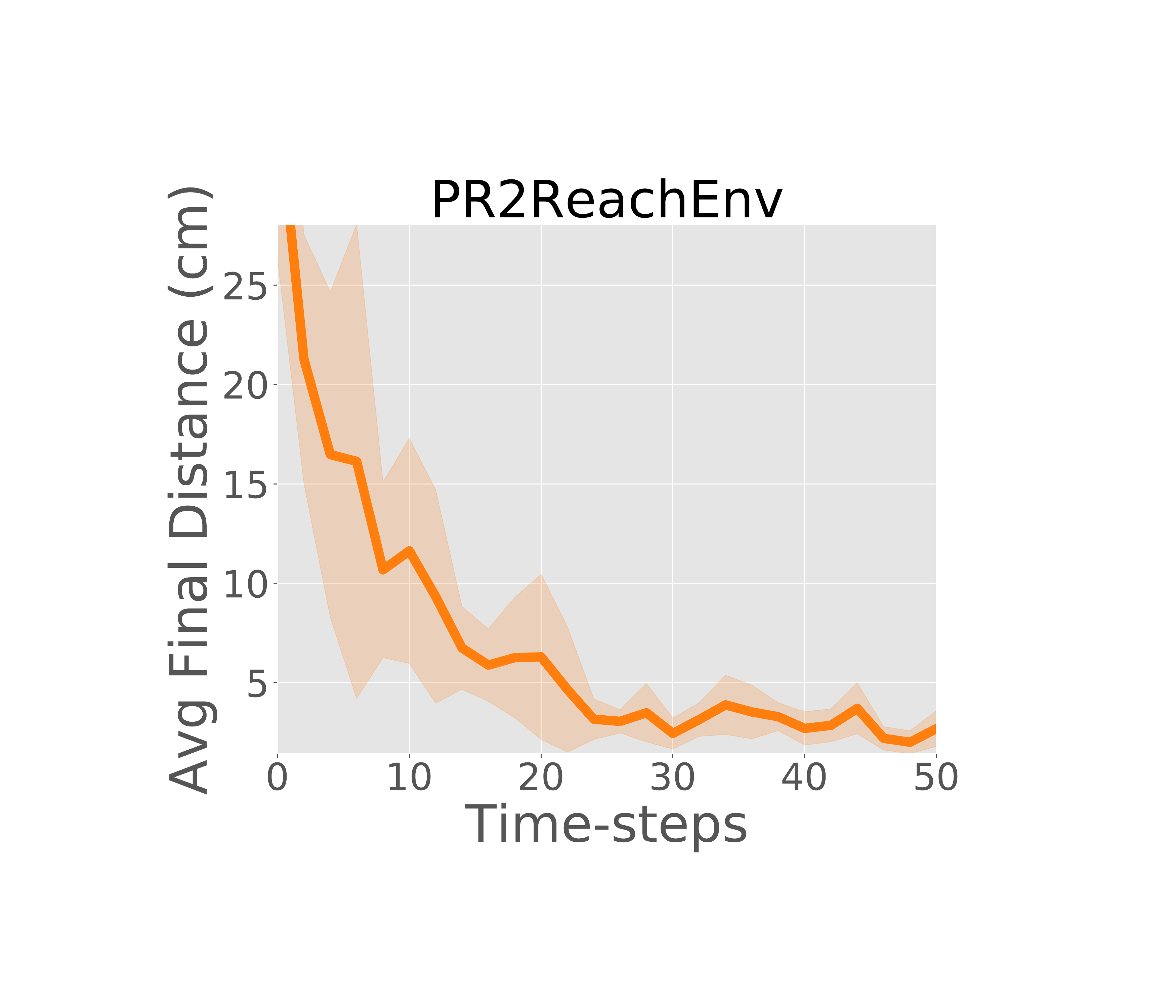}
        \centering
        \includegraphics[trim={3.7cm 2.5cm 5cm 2.5cm}, clip,  width=0.32\textwidth]{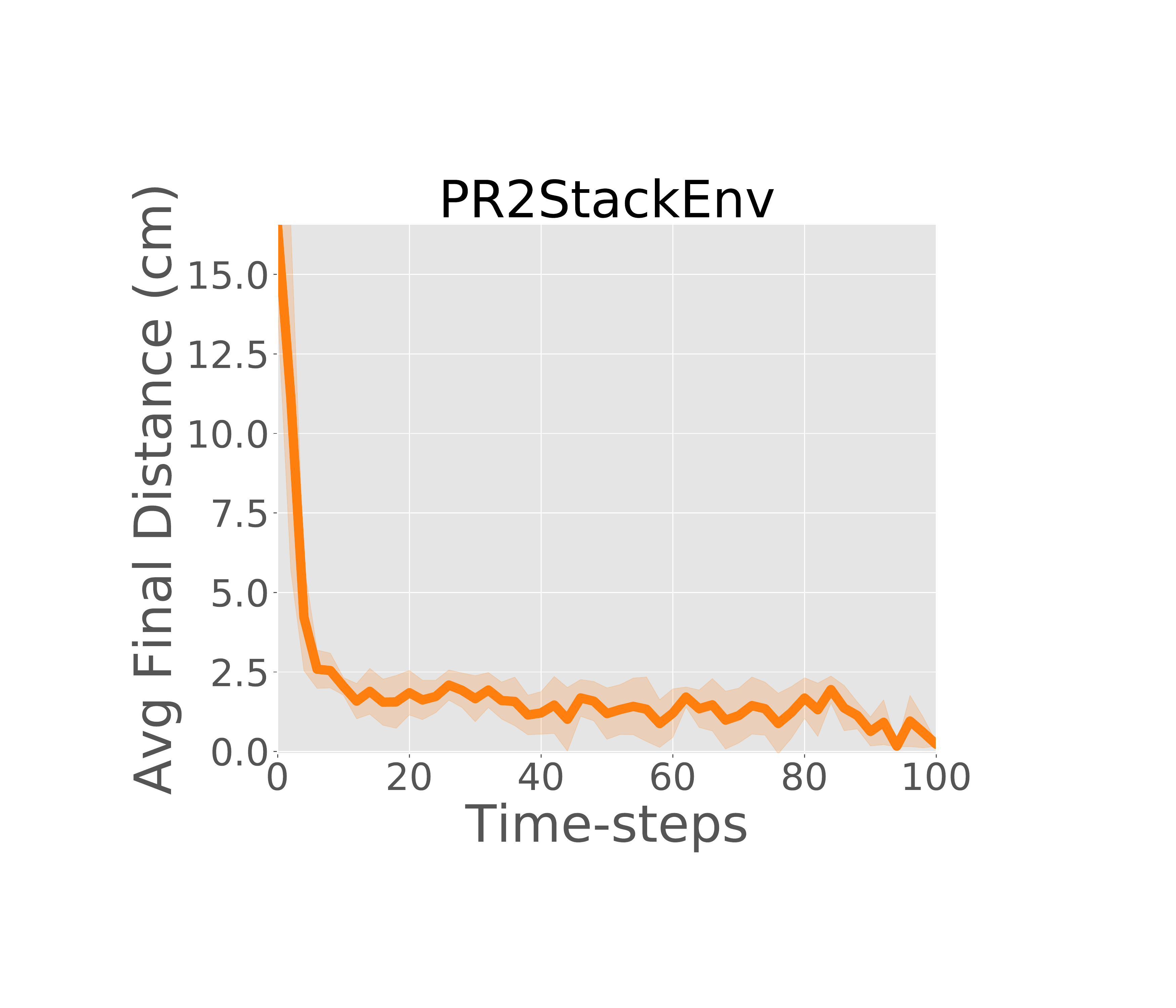}
        \caption{\small{Final distance attained in each of the tasks evaluated on the PR2 robot with asynch-MB-MPO.}}
    \label{fig:pr2_plots}
    \vspace{-0.5cm}
\end{figure*}

The results, shown in Figure~\ref{fig:pr2_plots}, show that asynchronous MB-MPO achives contact rich object manipulated tasks, such as lego stacking and shape matching, within 100 time-steps. This corresponds to 10 minutes of run time, matching similar speed performance attained in~\cite{levine2015learning}.  Videos of our method on the real robot environment, can be found at our website.\footnote{https://sites.google.com/view/asynch-mb-rl/home}

\section{Conclusion}
In this work we proposed an asynchronous framework for model-based reinforcement learning. Our empirical investigation shows that asynchronous model-based RL learns substantially faster than prior approaches. We characterized the key traits of asynchronous training that improves sample efficiency: policy regularization by interleaving policy learning and model learning, and better data collection by interleaving policy learning and data collection. Finally, we showcase the performance of asynchronous learning in real robotic manipulation, achieving to learn contact rich tasks within 10 min of run time. 
We release our code which supports an arbitrary number of data, model or policy workers and could be run across machines. 
An enticing direction for future work is the application of asynchronous learning to more complex real robotics tasks. 


\clearpage


\small{

}  


\begin{thebibliography}{39}
\providecommand{\natexlab}[1]{#1}
\providecommand{\url}[1]{\texttt{#1}}
\expandafter\ifx\csname urlstyle\endcsname\relax
  \providecommand{\doi}[1]{doi: #1}\else
  \providecommand{\doi}{doi: \begingroup \urlstyle{rm}\Url}\fi

\bibitem[Kaelbling et~al.(1996)Kaelbling, Littman, and
  Moore]{kaelbling1996survey}
L.~P. Kaelbling, M.~L. Littman, and A.~W. Moore.
\newblock Reinforcement learning: {A} survey.
\newblock \emph{CoRR}, cs.AI/9605103, 1996.

\bibitem[Janner et~al.(2019)Janner, Fu, Zhang, and Levine]{janner2019mbpo}
M.~Janner, J.~Fu, M.~Zhang, and S.~Levine.
\newblock When to trust your model: Model-based policy optimization.
\newblock \emph{CoRR}, abs/1906.08253, 2019.

\bibitem[Chua et~al.(2018)Chua, Calandra, McAllister, and Levine]{chua2018deep}
K.~Chua, R.~Calandra, R.~McAllister, and S.~Levine.
\newblock Deep reinforcement learning in a handful of trials using
  probabilistic dynamics models.
\newblock \emph{arXiv preprint arXiv:1805.12114}, 2018.

\bibitem[Clavera et~al.(2018)Clavera, Rothfuss, Schulman, Fujita, Asfour, and
  Abbeel]{clavera2018mbmpo}
I.~Clavera, J.~Rothfuss, J.~Schulman, Y.~Fujita, T.~Asfour, and P.~Abbeel.
\newblock Model-based reinforcement learning via meta-policy optimization.
\newblock \emph{CoRR}, abs/1809.05214, 2018.

\bibitem[Buckman et~al.(2018)Buckman, Hafner, Tucker, Brevdo, and
  Lee]{buckman2018steve}
J.~Buckman, D.~Hafner, G.~Tucker, E.~Brevdo, and H.~Lee.
\newblock Sample-efficient reinforcement learning with stochastic ensemble
  value expansion.
\newblock \emph{CoRR}, abs/1807.01675, 2018.
\newblock URL \url{http://arxiv.org/abs/1807.01675}.

\bibitem[Wang et~al.(2019)Wang, Bao, Clavera, Hoang, Wen, Langlois, Zhang,
  Zhang, Abbeel, and Ba]{wang2019mbbench}
T.~Wang, X.~Bao, I.~Clavera, J.~Hoang, Y.~Wen, E.~Langlois, S.~Zhang, G.~Zhang,
  P.~Abbeel, and J.~Ba.
\newblock Benchmarking model-based reinforcement learning.
\newblock \emph{CoRR}, abs/1907.02057, 2019.

\bibitem[Luo et~al.(2019)Luo, Xu, Li, Tian, Darrell, and
  Ma]{luo2018algorithmic}
Y.~Luo, H.~Xu, Y.~Li, Y.~Tian, T.~Darrell, and T.~Ma.
\newblock Algorithmic framework for model-based deep reinforcement learning
  with theoretical guarantees.
\newblock \emph{ICLR}, 2019.

\bibitem[Todorov et~al.(2012)Todorov, Erez, and Tassa]{2012mujoco}
E.~Todorov, T.~Erez, and Y.~Tassa.
\newblock Mujoco: A physics engine for model-based control.
\newblock In \emph{Intelligent Robots and Systems (IROS), 2012 IEEE/RSJ
  International Conference on}, pages 5026--5033. IEEE, 2012.

\bibitem[Levine et~al.(2015)Levine, Wagener, and Abbeel]{levine2015learning}
S.~Levine, N.~Wagener, and P.~Abbeel.
\newblock Learning contact-rich manipulation skills with guided policy search.
\newblock In \emph{2015 IEEE international conference on robotics and
  automation (ICRA)}, pages 156--163. IEEE, 2015.

\bibitem[Kurutach et~al.(2018)Kurutach, Clavera, Duan, Tamar, and
  Abbeel]{kurutach2018model}
T.~Kurutach, I.~Clavera, Y.~Duan, A.~Tamar, and P.~Abbeel.
\newblock Model-ensemble trust-region policy optimization.
\newblock \emph{arXiv preprint arXiv:1802.10592}, 2018.

\bibitem[Sutton(1991{\natexlab{a}})]{sutton1991dyna}
R.~S. Sutton.
\newblock Dyna, an integrated architecture for learning, planning, and
  reacting.
\newblock \emph{ACM SIGART Bulletin}, 2\penalty0 (4):\penalty0 160--163,
  1991{\natexlab{a}}.

\bibitem[Sutton(1991{\natexlab{b}})]{sutton1991planning}
R.~S. Sutton.
\newblock Planning by incremental dynamic programming.
\newblock In \emph{Machine Learning Proceedings 1991}, pages 353--357.
  Elsevier, 1991{\natexlab{b}}.

\bibitem[Nagabandi et~al.(2017)Nagabandi, Kahn, Fearing, and
  Levine]{nagabandi2017neural}
A.~Nagabandi, G.~Kahn, R.~S. Fearing, and S.~Levine.
\newblock Neural network dynamics for model-based deep reinforcement learning
  with model-free fine-tuning.
\newblock \emph{arXiv preprint arXiv:1708.02596}, 2017.

\bibitem[Deisenroth and Rasmussen(2011)]{deisenroth2011pilco}
M.~Deisenroth and C.~E. Rasmussen.
\newblock Pilco: A model-based and data-efficient approach to policy search.
\newblock In \emph{Proceedings of the 28th International Conference on machine
  learning (ICML-11)}, pages 465--472, 2011.

\bibitem[Heess et~al.(2015)Heess, Wayne, Silver, Lillicrap, Erez, and
  Tassa]{heess2015learning}
N.~Heess, G.~Wayne, D.~Silver, T.~Lillicrap, T.~Erez, and Y.~Tassa.
\newblock Learning continuous control policies by stochastic value gradients.
\newblock In \emph{Advances in Neural Information Processing Systems}, pages
  2944--2952, 2015.

\bibitem[Tassa et~al.(2012)Tassa, Erez, and Todorov]{tassa2012synthesis}
Y.~Tassa, T.~Erez, and E.~Todorov.
\newblock Synthesis and stabilization of complex behaviors through online
  trajectory optimization.
\newblock In \emph{Intelligent Robots and Systems (IROS), 2012 IEEE/RSJ
  International Conference on}, pages 4906--4913. IEEE, 2012.

\bibitem[Levine and Koltun(2013)]{levine2013guided}
S.~Levine and V.~Koltun.
\newblock Guided policy search.
\newblock In \emph{International Conference on Machine Learning}, pages 1--9,
  2013.

\bibitem[Recht et~al.(2011)Recht, Re, Wright, and Niu]{recht2011hogwild}
B.~Recht, C.~Re, S.~Wright, and F.~Niu.
\newblock Hogwild: A lock-free approach to parallelizing stochastic gradient
  descent.
\newblock In J.~Shawe-Taylor, R.~S. Zemel, P.~L. Bartlett, F.~Pereira, and
  K.~Q. Weinberger, editors, \emph{Advances in Neural Information Processing
  Systems 24}, pages 693--701. Curran Associates, Inc., 2011.

\bibitem[Dean et~al.(2012)Dean, Corrado, Monga, Chen, Devin, Mao, aurelio
  Ranzato, Senior, Tucker, Yang, Le, and Ng]{dean2012largescale}
J.~Dean, G.~Corrado, R.~Monga, K.~Chen, M.~Devin, M.~Mao, M.~aurelio Ranzato,
  A.~Senior, P.~Tucker, K.~Yang, Q.~V. Le, and A.~Y. Ng.
\newblock Large scale distributed deep networks.
\newblock In F.~Pereira, C.~J.~C. Burges, L.~Bottou, and K.~Q. Weinberger,
  editors, \emph{Advances in Neural Information Processing Systems 25}, pages
  1223--1231. Curran Associates, Inc., 2012.

\bibitem[Nair et~al.(2015)Nair, Srinivasan, Blackwell, Alcicek, Fearon, Maria,
  Panneershelvam, Suleyman, Beattie, Petersen, Legg, Mnih, Kavukcuoglu, and
  Silver]{nair2015gorila}
A.~Nair, P.~Srinivasan, S.~Blackwell, C.~Alcicek, R.~Fearon, A.~D. Maria,
  V.~Panneershelvam, M.~Suleyman, C.~Beattie, S.~Petersen, S.~Legg, V.~Mnih,
  K.~Kavukcuoglu, and D.~Silver.
\newblock Massively parallel methods for deep reinforcement learning.
\newblock \emph{CoRR}, abs/1507.04296, 2015.

\bibitem[Babaeizadeh et~al.(2016)Babaeizadeh, Frosio, Tyree, Clemons, and
  Kautz]{Babaeizadeh2016ga3c}
M.~Babaeizadeh, I.~Frosio, S.~Tyree, J.~Clemons, and J.~Kautz.
\newblock {GA3C:} gpu-based {A3C} for deep reinforcement learning.
\newblock \emph{CoRR}, abs/1611.06256, 2016.

\bibitem[Espeholt et~al.(2018)Espeholt, Soyer, Munos, Simonyan, Mnih, Ward,
  Doron, Firoiu, Harley, Dunning, Legg, and Kavukcuoglu]{espeholt2018impala}
L.~Espeholt, H.~Soyer, R.~Munos, K.~Simonyan, V.~Mnih, T.~Ward, Y.~Doron,
  V.~Firoiu, T.~Harley, I.~Dunning, S.~Legg, and K.~Kavukcuoglu.
\newblock {IMPALA:} scalable distributed deep-rl with importance weighted
  actor-learner architectures.
\newblock \emph{CoRR}, abs/1802.01561, 2018.

\bibitem[Heess et~al.(2017)Heess, TB, Sriram, Lemmon, Merel, Wayne, Tassa,
  Erez, Wang, Eslami, Riedmiller, and Silver]{hess2017dppo}
N.~Heess, D.~TB, S.~Sriram, J.~Lemmon, J.~Merel, G.~Wayne, Y.~Tassa, T.~Erez,
  Z.~Wang, S.~M.~A. Eslami, M.~A. Riedmiller, and D.~Silver.
\newblock Emergence of locomotion behaviours in rich environments.
\newblock \emph{CoRR}, abs/1707.02286, 2017.
\newblock URL \url{http://arxiv.org/abs/1707.02286}.

\bibitem[Mnih et~al.(2016)Mnih, Badia, Mirza, Graves, Lillicrap, Harley,
  Silver, and Kavukcuoglu]{mnih2016asynchronous}
V.~Mnih, A.~P. Badia, M.~Mirza, A.~Graves, T.~Lillicrap, T.~Harley, D.~Silver,
  and K.~Kavukcuoglu.
\newblock Asynchronous methods for deep reinforcement learning.
\newblock In \emph{International Conference on Machine Learning}, pages
  1928--1937, 2016.

\bibitem[Stooke and Abbeel(2018)]{stooke2018accelrl}
A.~Stooke and P.~Abbeel.
\newblock Accelerated methods for deep reinforcement learning.
\newblock \emph{CoRR}, abs/1803.02811, 2018.

\bibitem[Mnih et~al.(2015)Mnih, Kavukcuoglu, Silver, Rusu, Veness, Bellemare,
  Graves, Riedmiller, Fidjeland, Ostrovski, et~al.]{mnih2015human}
V.~Mnih, K.~Kavukcuoglu, D.~Silver, A.~A. Rusu, J.~Veness, M.~G. Bellemare,
  A.~Graves, M.~Riedmiller, A.~K. Fidjeland, G.~Ostrovski, et~al.
\newblock Human-level control through deep reinforcement learning.
\newblock \emph{Nature}, 518\penalty0 (7540):\penalty0 529--533, 2015.

\bibitem[Schulman et~al.(2017)Schulman, Wolski, Dhariwal, Radford, and
  Klimov]{ppo}
J.~Schulman, F.~Wolski, P.~Dhariwal, A.~Radford, and O.~Klimov.
\newblock Proximal policy optimization algorithms.
\newblock \emph{arXiv preprint arXiv:1707.06347}, 2017.

\bibitem[Precup et~al.(2000)Precup, Sutton, and Singh]{precup2000eltraces}
D.~Precup, R.~S. Sutton, and S.~P. Singh.
\newblock Eligibility traces for off-policy policy evaluation.
\newblock In \emph{Proceedings of the Seventeenth International Conference on
  Machine Learning}, ICML '00, pages 759--766, San Francisco, CA, USA, 2000.
  Morgan Kaufmann Publishers Inc.
\newblock ISBN 1-55860-707-2.

\bibitem[Precup et~al.(2001)Precup, Sutton, and Dasgupta]{precup2001tdlearning}
D.~Precup, R.~S. Sutton, and S.~Dasgupta.
\newblock Off-policy temporal difference learning with function approximation.
\newblock In \emph{Proceedings of the Eighteenth International Conference on
  Machine Learning}, ICML '01, pages 417--424, San Francisco, CA, USA, 2001.
  Morgan Kaufmann Publishers Inc.
\newblock ISBN 1-55860-778-1.
\newblock URL \url{http://dl.acm.org/citation.cfm?id=645530.655817}.

\bibitem[Gu et~al.(2016)Gu, Holly, Lillicrap, and Levine]{gu2016asynchrobotics}
S.~Gu, E.~Holly, T.~P. Lillicrap, and S.~Levine.
\newblock Deep reinforcement learning for robotic manipulation.
\newblock \emph{CoRR}, abs/1610.00633, 2016.

\bibitem[{Moldovan} et~al.(2015){Moldovan}, {Levine}, {Jordan}, and
  {Abbeel}]{moldovan2015optimsm}
T.~M. {Moldovan}, S.~{Levine}, M.~I. {Jordan}, and P.~{Abbeel}.
\newblock Optimism-driven exploration for nonlinear systems.
\newblock In \emph{2015 IEEE International Conference on Robotics and
  Automation (ICRA)}, pages 3239--3246, May 2015.
\newblock \doi{10.1109/ICRA.2015.7139645}.

\bibitem[Lioutikov et~al.(2014)Lioutikov, Paraschos, Peters, and
  Neumann]{lioutikov2014sample}
R.~Lioutikov, A.~Paraschos, J.~Peters, and G.~Neumann.
\newblock Sample-based informationl-theoretic stochastic optimal control.
\newblock In \emph{2014 IEEE International Conference on Robotics and
  Automation (ICRA)}, pages 3896--3902. IEEE, 2014.

\bibitem[Clavera et~al.(2018)Clavera, Nagabandi, Fearing, Abbeel, Levine, and
  Finn]{nagabandi2018l2a}
I.~Clavera, A.~Nagabandi, R.~S. Fearing, P.~Abbeel, S.~Levine, and C.~Finn.
\newblock Learning to adapt: Meta-learning for model-based control.
\newblock \emph{CoRR}, abs/1803.11347, 2018.

\bibitem[Haarnoja et~al.(2018{\natexlab{a}})Haarnoja, Zhou, Hartikainen,
  Tucker, Ha, Tan, Kumar, Zhu, Gupta, Abbeel, and Levine]{haarnoja2018sacapp}
T.~Haarnoja, A.~Zhou, K.~Hartikainen, G.~Tucker, S.~Ha, J.~Tan, V.~Kumar,
  H.~Zhu, A.~Gupta, P.~Abbeel, and S.~Levine.
\newblock Soft actor-critic algorithms and applications.
\newblock \emph{CoRR}, abs/1812.05905, 2018{\natexlab{a}}.

\bibitem[Haarnoja et~al.(2018{\natexlab{b}})Haarnoja, Zhou, Ha, Tan, Tucker,
  and Levine]{haaronja2018l2walk}
T.~Haarnoja, A.~Zhou, S.~Ha, J.~Tan, G.~Tucker, and S.~Levine.
\newblock Learning to walk via deep reinforcement learning.
\newblock \emph{CoRR}, abs/1812.11103, 2018{\natexlab{b}}.

\bibitem[{Hafner} and {Riedmiller}(2007)]{hafner2007rlrobcup}
R.~{Hafner} and M.~{Riedmiller}.
\newblock Neural reinforcement learning controllers for a real robot
  application.
\newblock In \emph{Proceedings 2007 IEEE International Conference on Robotics
  and Automation}, pages 2098--2103, April 2007.
\newblock \doi{10.1109/ROBOT.2007.363631}.

\bibitem[Gullapalli et~al.(1992)Gullapalli, Grupen, and
  Barto]{gullapalli1992learning}
V.~Gullapalli, R.~A. Grupen, and A.~G. Barto.
\newblock Learning reactive admittance control.
\newblock In \emph{Proceedings 1992 IEEE International Conference on Robotics
  and Automation}, pages 1475--1480. IEEE, 1992.

\bibitem[Brockman et~al.(2016)Brockman, Cheung, Pettersson, Schneider,
  Schulman, Tang, and Zaremba]{gym}
G.~Brockman, V.~Cheung, L.~Pettersson, J.~Schneider, J.~Schulman, J.~Tang, and
  W.~Zaremba.
\newblock Openai gym.
\newblock \emph{arXiv preprint arXiv:1606.01540}, 2016.

\bibitem[Schulman et~al.(2015)Schulman, Levine, Abbeel, Jordan, and
  Moritz]{trpo}
J.~Schulman, S.~Levine, P.~Abbeel, M.~Jordan, and P.~Moritz.
\newblock Trust region policy optimization.
\newblock In \emph{Proceedings of the 32nd International Conference on Machine
  Learning (ICML-15)}, pages 1889--1897, 2015.

\end{thebibliography}
\end{document}